# Fine-tuning Pretrained Multilingual BERT Model for Indonesian Aspect-based Sentiment Analysis

Annisa Nurul Azhar
*School of Electrical Engineering and Informatics*
*Institut Teknologi Bandung*
Bandung, Indonesia
23519025@std.stei.itb.ac.id

Masayu Leylia Khodra
*School of Electrical Engineering and Informatics*
*Institut Teknologi Bandung*
Bandung, Indonesia
masayu@stei.itb.ac.id

*Abstract*—Although previous research on Aspect-based Sentiment Analysis (ABSA) for Indonesian reviews in hotel domain has been conducted using CNN and XGBoost, its model did not generalize well in test data and high number of OOV words contributed to misclassification cases. Nowadays, most state-of-the-art results for wide array of NLP tasks are achieved by utilizing pretrained language representation. In this paper, we intend to incorporate one of the foremost language representation model, BERT, to perform ABSA in Indonesian reviews dataset. By combining multilingual BERT (m-BERT) with task transformation method, we manage to achieve significant improvement by 8% on the F1-score compared to the result from our previous study.

*Keywords—pretrained language model, aspect-based sentiment analysis, sentence-pair classification, sequential transfer learning, fine-tuning*

## I. INTRODUCTION

Sentiment analysis has been widely used among companies to automatically extract opinions about their products or services. Several companies may need more fine-grained analysis by employing aspect-based sentiment analysis (ABSA). Typically, ABSA task is broken down into two subtask, aspect extraction/categorization and sentiment classification (Liu, B., 2012). Aspect extraction/categorization aims to extract aspect terms (aspect extraction) e.g. "helpful staff" or categorize aspect terms (aspect categorization) into some predetermined categories e.g. "service". Next, the process of determining sentiment e.g. "positive", "negative" or "neutral" towards each aspect is conducted in sentiment classification subtask. As a final result, aspect-based sentiment analysis system returns output pairs of aspect and its sentiment which exist in the review text.

Previous study about ABSA for Indonesian reviews has been conducted by utilizing Convolutional Neural Network (CNN) and Extreme Gradient Boosting (XGBoost) by Azhar A.N. (2019) and Azhar, et al. (2019) to perform both aspect categorization and sentiment classification. CNN is employed as a feature extractor and XGBoost for top-level classifier. However, it was found that the model does not generalize well in the test set. Misclassification cases due to out-of-vocabulary (OOV) words still occur a lot. Therefore, new technique needs to be utilized to build a better model.

Transfer learning from pretrained language representation model seems to be a promising approach to solve the problems mentioned above. Pretrained models are commonly pretrained on a large-scale dataset hence it would generalize well and most of words would present in the vocabulary so the number of OOV words could be suppressed. Moreover, these days, techniques with transfer learning approach has been successfully achieving higher performance on various natural language processing tasks such as question answering, sentiment analysis, and semantic textual similarity, as compared to non-transfer learning techniques. This progress certainly affected by the evolution of pretraining method used to pretrain the pretrained model. One of the foremost methods is Bidirectional Encoder Representation from Transformers (BERT) by Devlin et al. (2018) which able to achieve state-of-the-art result on most of natural language processing tasks. BERT is the first method to introduce unsupervised and deeply bidirectional approach for pretraining a language representation model.

Pretrained BERT model has been extensively used to address ABSA task. Generally, there are two approaches for utilizing pretrained BERT model or any pretrained model for specific NLP task according to recent work by Ruder, S. (2019) on transfer learning in NLP namely feature extraction and fine-tuning. The difference between the two approaches is that in feature extraction, parameters of the pretrained model are freezed meanwhile in fine-tuning they are adjusted to be more specific to the downstream task. For most of the time, fine-tuning approach works better than the feature extraction. It makes sense because fine-tuning approach is able to obtain task-specific knowledge. One of the recent works which fine-tuned the pretrained BERT model for Targeted ABSA (T-ABSA) task is conducted by Sun, et al. (2019) and it is able to achieve state-of-the-art result on Sentihood and Semeval-2014 dataset.

ABSA itself is considered as a single sentence classification task because it takes a single sentence as an input and perform classification on it. According to the original work on BERT by Devlin et al. (2018), the pretrained BERT model could also be used for single sentence classification tasks despite pretrained for sentence-pair classification task which is Next Sentence Prediction (NSP). However, Sun, et al. (2019) found that there is no significant raise on T-ABSA performance by directly fine-tuned the pretrained model. So, they proposed to modify T-ABSA task first from single sentence classification task into sentence-pair classification task before fine-tuning the pretrained model by constructing auxiliary sentences to be paired with the original review texts. This approach then managed to obtain significantly higher performance than the single sentence method.

In this work, we conduct a comparative experiment to search for the best combination of strategy for using pretrained model (feature extraction and fine-tuning) and approach to solve ABSA task (as single sentence classification task and as sentence pair classification task) in Indonesian language. We will use the proposed method by Sun, et al. (2019) to construct the auxiliary sentences to convert ABSA into sentence-pair classification task. Since pretrained BERT model for Indonesian language is currently non-existent for public use, we will use the multilingual one which is already available for public. Next, we perform hyperparameter tuning while training the model using the best combination from the previous experiment. We also analyze the difference between using a BERT-based unilingual pretrained model and multilingual one in terms of vocabulary issue.

This paper is organized as follows. Section 2 discusses related works i.e. sequential transfer learning (Ruder, S., 2019), m-BERT (Devlin, et al., 2018), and sentence-pair classification for ABSA (Sun, et al., 2019). Our methodology will be discussed in section 3. The implementation and dataset used for this work will be described in section 4. Section 5 and section 6 will explain about the experiments, result, and the analysis. Finally, conclusions of this work will be discussed in section 7.

## II. Related Work

### A. Sequential Transfer Learning

The term "sequential transfer learning" is first introduced by Ruder, S. (2019) to define settings in transfer learning when source and target tasks are not the same and the trainings are done sequentially. There are two main steps in sequential transfer learning, namely the pretraining step and the adaptation step. By using a BERT pretrained model, we do not need to perform the first step since it is already done before. We only need to do the adaptation step to transfer the knowledge into a target task. For the adaptation step, there are generally two strategies, namely feature extraction and fine-tuning. In the feature extraction strategy, representations generated from the learning of the pretrained model are used as input features for training a separate model. The parameters in the pretrained model will be frozen so that they will not change during training on the target task. Meanwhile, in the fine-tuning strategy, learning for the target task is done directly on the pretrained model. The parameters of the pretrained model then will be adjusted during the training process on the target task (Ruder, S., 2019).

### B. Multilingual BERT

Bidirectional Encoder Representations from Transformers or BERT is a language model that utilizes the encoder mechanism on Transformers in doing task language modeling. The term bidirectional contained in BERT explains that in pretraining phase, the encoder in BERT reads the entire sequence of input words all at once (bidirectional), different from the techniques applied to previous language models that read the sequence of input words sequentially from left to right or from right to left only (unidirectional). This resulted in the model being built to be able to understand the context of the words before (left-context) and the context of the words after (right context) of a word. With the ability to understand the context, the BERT instructional language model can improve performance in various tasks in NLP such as Natural Language Inference (NLI) and Question Answering (QA) through fine-tuning method by only adding one classification layer (Devlin, et al., 2018).

The architecture of the BERT model is a multi-layer bidirectional Transformer encoder as implemented by Vaswani, et al. (2017). In their work, Devlin, et al. (2018) states the number of layers which are also Transformer blocks as L, hidden size as H, and self-attention heads as A. There are two variations of the BERT model based on its size, namely BERT-Base with $L = 12$, $H = 768$, and $A = 12$ and BERT-Large with $L = 24$, $H = 1024$, and $A = 16$).

BERT pretrained multilingual language model (m-BERT) is pretrained with training set in which there are 104 languages, including Indonesian language. Indonesian training data is taken from all Wikipedia dumps in Indonesian, which consists of 532,806 articles. For the pretraining process, BERT is trained to complete two tasks namely masked language model (MLM) and next sentence prediction (NSP). In order to conduct bidirectional model training, some input tokens will be masked (replacing tokens with [MASK] tokens) randomly then the model will predict these tokens based on other tokens that are not masked. The prediction is done by adding one classification layer above the output of the encoder then the resulting output vector will be multiplied by the embedding matrix. Finally, the overall probability of words in the vocabulary is calculated by the softmax activation function. In its implementation, BERT changes 15% of the total input tokens to [MASK] tokens.

As for NSP, the BERT model needs to be trained with NSP task because some tasks in NLP such as NLI and QA require information about the relationship between two sentences (sentence-pair) but that information cannot be obtained by training in MLM task only. This process is done by providing input in the form of sentence pairs (sentence A and sentence B) then the model will predict whether sentence B is the next sentence after sentence A on the corpus. As many as 50% of the training data uses the actual follow-up sentence as sentence B and is labeled IsNext and the rest uses random sentences from the corpus and is labeled with NotNext. To be able to distinguish sentences 1 and sentences 2, BERT will perform several pre-processes of the input so that it becomes an appropriate input representation for the BERT model.

### C. ABSA as Sentence-pair Classification

Pre-trained BERT has been used in recent work conducted by Sun et al. (2019) to complete the ABSA task and successfully achieve state-of-the-art results in Semeval 2014 Task 4. The proposed technique is to construct auxiliary sentences from the initial sentence before being used as input for the BERT model. The basic idea is to change the ABSA task into a sentence pair classification task. There are four methods considered to generate auxiliary sentences namely Sentences for QA-M, Sentences for NLI-M, Sentences for QA-B, and Sentences for NLI-B. Auxiliary sentences will contain all possible combinations of entity-aspect-sentiment pairs (in 'B' settings) or entity-aspect pairs (in 'M' settings). Then, in order to complete the ABSA task, fine-tuning is done by adding a classification layer to the output of the Transformer.

QA-M and QA-B methods transform T-ABSA task into question-answering task therefore the auxiliary sentences used in those methods are in question sentence form. Meanwhile, NLI-B and NLI-M methods transform the task into natural language inference task and use pseudo-sentence form for the auxiliary sentences. For 'M' and 'B', 'M' stands for multiclass and 'B' for binary. The dataset transformed using QA-M and NLI-M method will have its sentiment classes as the labels. Generally, there are two classes for sentiment polarity namely 'positive' and 'negative'. However, 'none' class is needed to point out pairs of entity-aspect contained in generated auxiliary sentences which are not present in the actual review text so the sentiment polarities cannot be classified as 'positive' or 'negative'. Since it will always be more than 2 classes in the dataset and each instance should only belong to one class, therefore it is considered as multiclass setting. As for QA-B and NLI-B methods, the labels will either be '0' or '1' therefore it is considered as binary settings. Label '0' means the combinations of entity-aspect-sentiment in the corresponding auxiliary sentence is false since it is not present in the actual review text and vice versa for label '1'.

III. METHODOLOGY

In this section, we will describe our methodology. We will divide it into two parts. The first one will focus on our methodology to build ABSA model by using single sentence classification approach and the second one will be for the one which uses sentence pair classification approach.

Methodology to build ABSA model by using m-BERT pretrained model based on single sentence classification approach is shown in Fig. 1. We train one model to perform aspect categorization which is a multilabel classification task with one-vs-all strategy (Bishop, 2006). For sentiment classification, we train a single sentiment classification model for each aspect category. There are 10 aspect categories in our dataset so in total, we train 11 models to solve ABSA task. Detailed explanation for each category are already described in our previous work (Azhar, et al., 2019).

The processes for constructing the models are as follows. First, we collect our dataset then create new smaller datasets from the original dataset by grouping each review text based on what aspect categories that it has into a new dataset for training the sentiment classification models. After that, we train the models separately. However, before the training phase, we run some preprocessing steps; tokenization and feature extraction. Since we will use m-BERT pretrained model, for tokenization we also use BERT Tokenizer.

The difference between traditional tokenization process and BERT tokenization process is that BERT tokenization does not only perform basic tokenization steps (lower-casing, whitespace removal, etc.) but also uses WordPiece (Wu, et al., 2016; Schuster & Nakajima, 2012) tokenizer to tokenize the text. WordPiece is one of word segmentation techniques which will broke down words that are non-existing in the vocabulary into sub-words thus lowering the number of OOV words. Algorithm used to construct the sub-words is greedy first-longest-match. For example, given a sentence "kamarnya bersih dan pelayanannya bagus" (*room is clean and service is good*). BERT will tokenize the text into a list of sub-words ["kama", "##rny", "##a", "be", "##rsi", "##h", "dan", "pela", "##yanan", "##nya", "bag", "##us"]. As we can see from the tokenization result, base words in bahasa such as "kamar" (*room*), "bersih" (*clean*), dan "bagus" (*good*) are also splitted into sub-words. This happens because those words are not present in the vocabulary thus the model take the first-longest-match and split each word into sub-words based on it.

A. *Methodology for Single Sentence Classification*

In feature extraction step we actually convert the tokenized input into BERT input representation. There are three components of the input representation namely Token Embeddings, Segment Embeddings, dan Position Embeddings. First, we insert [CLS] token as the first token and [SEP] token as the last token after the last sub-word from the tokenized input. Next, we lookup the embedding matrix to get embeddings for each token called Token Embeddings. Since BERT also accept sentence-pair input, Segment Embeddings are used to differentiate between token from the first sentence and from the second sentence. Token from the first sentence will be denoted by number "0" meanwhile token from the second sentence will be denoted by number "1". Position embeddings is just the position of each token in the entire input sequence. Finally, BERT input representation is the sum of Token Embeddings, Segment Embeddings, and Position Embeddings.

The next step is to train the m-BERT pretrained model on our dataset. To do that, we need to add a single classification layer on top of the last Transformer encoder. Weights of [CLS] token from the output of the last Transformer encoder will be fed as an input for the classification layer. Since we use one-vs-all strategy for aspect categorization, the number of nodes in the classification layer will be equal to the number of aspect categories and sigmoid activation function will be applied. As for sentiment classification, there is one output node on the classification layer with softmax activation function. In feature extraction strategy, only the classification layer is trained using the BERT output. Meanwhile, in fine-tuning strategy, the entire model is trained using the BERT output.

B. *Methodology for Sentence Pair Classification*

The methodology for sentence-pair classification is in fact pretty similar to the one for single sentence classification. The model construction process for sentence-pair classification approach can be seen in Fig. 2. However, in order to use sentence-pair classification approach, we need to construct auxiliary sentence first so we could pair them with the reviews in the original dataset.

There are four different methods to construct auxiliary sentences as described in Section II. We decided to use the NLI-B method straight-off due to limited resources to experiment with each method and also because it produces the highest result in most of the experiment scenarios. Thus, we generate all possible combinations of aspect category and sentiment polarity which are present in the original dataset then transform to pseudo-sentence form for the auxiliary sentences. With this construction method, train dataset will be composed by pairs of review sentence and auxiliary sentence which is a pseudo-sentence consists of pairs between aspect category and its sentiment polarity with binary label, "0" or "1". Label 0 means the aspect-sentiment pair is not present in the corresponding review text and label 1 means the opposite. The illustration of train dataset for sentence-pair classification can be seen in Table 1. The first sentence, which is the auxiliary sentence, is denoted as 'text_a', the actual review text is denoted as 'text_b', and the class is denoted as 'label'.

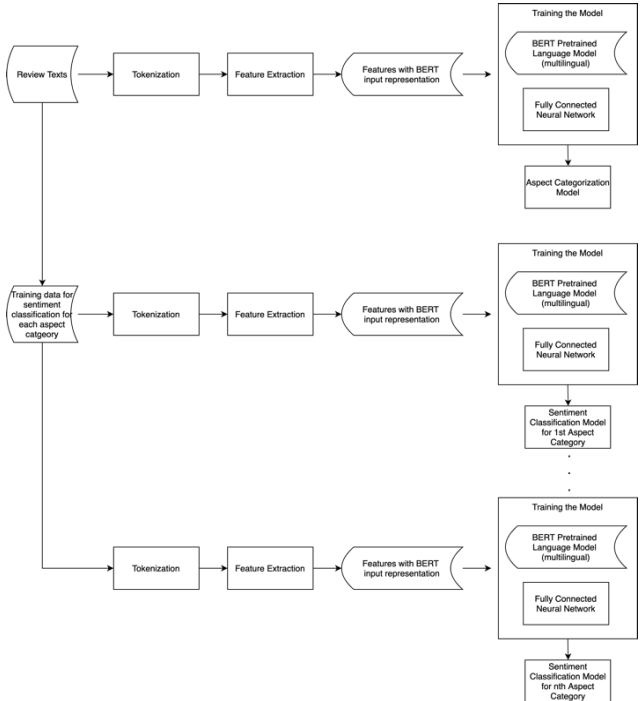

Fig. 1 Model Construction Process for Single Sentence Classification

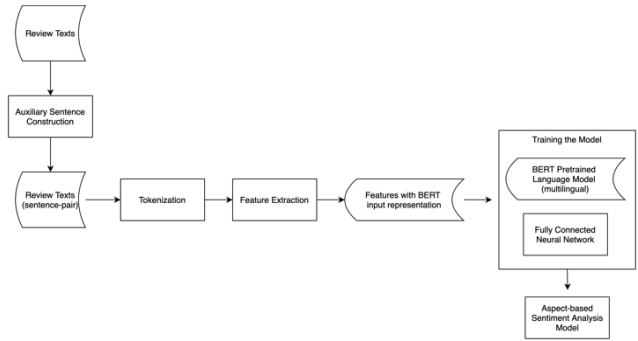

Fig. 2 Model Construction Process for Sentence-pair Classification

Table 1. Illustration of train dataset for sentence-pair classification.

| text_a (aux. sentence) | text_b (actual review text) | Label |
|---|---|---|
| service-positif (*service-positive*) | kamarnya bersih dan pelayanannya bagus (*the room is clean and the service is nice*) | 1 |
| service-negatif (*service-negative*) | kamarnya bersih dan pelayanannya bagus | 0 |
| service-none (*service-none*) | kamarnya bersih dan pelayanannya bagus | 0 |
| kebersihan-positif (*cleanliness-positive*) | kamarnya bersih dan pelayanannya bagus | 1 |
| kebersihan-negatif (*cleanliness-negative*) | kamarnya bersih dan pelayanannya bagus | 0 |
| kebersihan-none (*cleanliness-none*) | kamarnya bersih dan pelayanannya bagus | 0 |

The next steps, tokenization and feature extraction, are typically the same with the single-sentence-approach. The difference is we have to add 2 separator tokens [SEP], the first one placed between the last token of the first sentence and the first token of the second sentence. The other one is placed at the end of the second sentence after its last token. For training phase in sentence-pair classification approach, we only need to train one classifier to perform both aspect categorization and sentiment classification. Simply add one classification layer with on top of the Transformer output and apply softmax activation function. The architecture of the BERT model for sentence-pair classification can be seen in Fig 3.

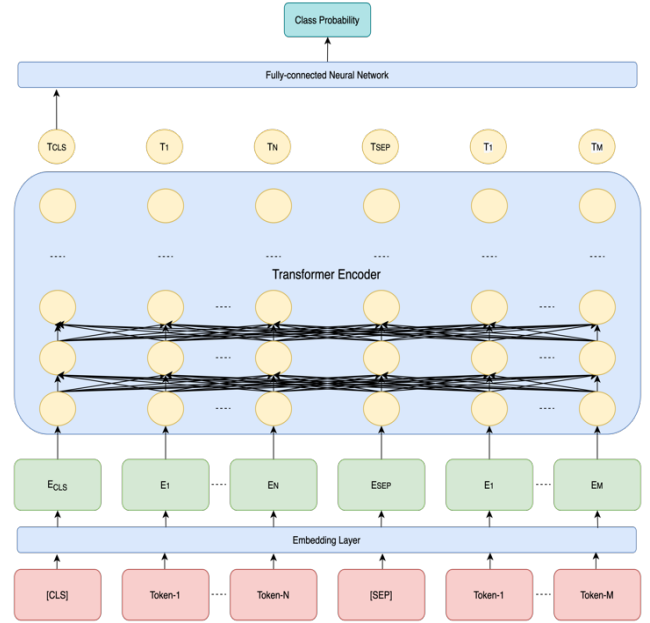

Fig. 3 BERT Model Architecture for Sentence-pair Classification

IV. IMPLEMENTATION AND DATASET

The implementation of this work uses Python programming language with the Tensorflow 1.15 library and is run at Google Collaboratory (https://colab.research.google.com/) using the Tensor Processing Unit or TPU hardware accelerator (https://cloud.google.com / tpu) with 8 cores. Most of the code for preprocessing and utilizing the BERT pretrained language model was adapted from the code in the BERT repository (https://github.com/google-research/bert) and modified so it would be able to perform multilabel classifications for aspect categorization model, able to use feature extraction strategy, and could use distributed training with TPU. Script for generating sentence-pair training data and do performance evaluation is implemented by ourselves using the Numpy, Pandas, and Sklearn libraries.

For experiment, we use Airyrooms' (former Indonesian accommodation network orchestrator) users review dataset with a total of 9448 instances. As for dataset that has been modified into sentence-pairs, the number of instances has increased to around 283,483 instances. Furthermore, this two datasets will be split into 80% training data and 20% validation data. We will also use the same test data as used in previous studies (Azhar, A.N., 2019; Azhar, et al., 2019) so that the performance of the two models can be compared.

V. EXPERIMENT

As mentioned in the introduction regarding the purpose of the experiment, there are several factors that will be compared in the experiment, namely the problem solving approach, the strategy to adapt a pretrained language model, and hyperparameters value to choose when learning from the BERT pretrained language model (hyperparameter tuning). The details of the experimental factors can be seen in Table 2.

Table 2. Details of the experimental factors

| Task-solving approach | Single sentence classification, sentence-pair classification |
|---|---|
| Strategy for model adaptation | Feature extraction, fine-tuning |
| Learning hyperparameters | Learning rate, batch size |

Due to limited resource and time, we choose what hyperparameters we would like to tune based on fine-tuning procedure in Devlin, et al. (2018) except epoch, which are learning rate and batch size. There are 2 values for learning rate and 2 values for batch size that will be searched for the best combination using the grid search. The values of the parameters that will be combined can be seen in Table 3. The number of epochs is 25 epochs. The experimental scheme used is hold-out with 80% data as training data and 20% data as validation data.

Table 3. Details of hyperparameter tuning

| Hyperparameter | Values | Number of Combinations |
|---|---|---|
| Learning rate | 3e5, 2e-5 | 2 |
| Batch size | 16, 32 | 2 |
| | Total | 4 |

## VI. Result and Analysis

### A. Experiment Result for Task-solving approach and strategy for model adaptation

Experiment result for the first scenario can be seen in Table 4. By using the fine-tuning strategy in the two approaches, we could see better performances than the feature-extraction strategy with a difference of around 8%. The fine-tuning strategy obtained better result because the parameters in the pretrained language model are also updated so that it is exposed to the specific knowledge of the target task. Meanwhile, in the feature extraction strategy, the parameters in the pretrained language model are frozen so that only the parameters of the last layer (classification layer) are trained.

Table 4. Experiment result for Task-solving approach and strategy for model adaptation

| Task-solving approach | Strategy for model adaptation | F1-score |
|---|---|---|
| Single Sentence Classification | Feature Extraction | 0.8764 |
| | Fine-tuning | 0.8943 |
| Sentence-Pair Classification | Feature Extraction | 0.9553 |
| | Fine-tuning | 0.9751 |

It also can be seen from Table 4 that in general the sentence-pair classification approach gets significantly better performance compared to the single sentence classification approach, same as the overall experiment result done by Sun, et al. (2019). Therefore, it can be taken that the ability of the pretrained model can be exploited more optimally by converting single sentence classification task into sentence pair classification task so it will be more similar to NSP task which the pretrained model was trained on.

### B. Experiment Result for Hyperparameter Tuning

After finding the best task-solving approach and adaptation strategy in the previous experiment, another experiment is conducted to find the best combination of hyperparameter values combination using sentence-pair classification as task-solving approach and fine-tuning as the adaptation strategy. The results of the experiment can be seen in Table 5. The selection of the learning rate value greatly determines the performance of the model because in the fine-tuning strategy, updates made to the parameters of the pretrained model depend on the value of the learning rate. If the value of learning rate is set too large, it is feared that the adjustments made will be excessive so that it can lose the knowledge that has been obtained by the pretrained model after going through the pretraining process. This can cause the model to overfit.

Based on Table 5, the best combination of learning rate and batch size values is 2e-5 and 32. It can be seen that the selection of a smaller learning rate at 2e-5 obtain slightly better result than 3e-5 so learning rate value of 2e-5 is chosen to be used for full training. However, the difference in performance of each combination is not significant. This is consistent with observations made in the work from Devlin, et al., 2018 that large datasets (more than 100,000 instances) are far less sensitive to hyperparameter selection than smaller datasets.

Table 5. Experiment result for Hyperparameter tuning

| Hyperparameter | Values | F1-score |
|---|---|---|
| Learning rate | 3e-5 | 0.9721 |
| Batch size | 16 | |
| Learning rate | 2e-5 | 0.9739 |
| Batch size | 16 | |
| Learning rate | 3e-5 | 0.9744 |
| Batch size | 32 | |
| Learning rate | 2e-5 | 0.9751 |
| Batch size | 32 | |

### C. Evaluation

After the final model was trained using the best configuration based on experiment results, the performance of the model was compared with the performance of the model from previous studies by Azhar, A.N. (2019) which used the CNN-XGBoost method to solve ABSA task. We use the same evaluation method to for both problem solving method. A performance comparison between the two models in test dataset can be seen in Table 6. In Table 6, it can be seen that by using problem solving method proposed in this work, the performance of the model in test dataset has significantly increased. This means that the generalization of the model is getting much better compared to previous studies. However, it should be remembered that the training time for fine-tuning the BERT pretrained model with auxiliary sentences has a considerable difference with the training time of the model with the CNN-XGBoost method. Even with TPU as hardware accelerator, the training time is 6 times longer than the CNN-XGBoost training time without hardware accelerator.

Table 6. Evaluation Result

| Problem Solving Method | F1-score |
|---|---|
| CNN-XGBoost | 0.8958 |
| *Fine-tuning* m-BERT + Auxiliary Sentences | **0.9765** |

### D. *Analysis*

Compared to misclassification cases that occur in evaluation from previous studies, many of them do not re-occur in current evaluation. These sentences are then classified correctly by the model built by fine-tuning the m-BERT pretrained model in this study. This is most likely caused by the improved language representation by BERT and the tokenization technique. BERT uses a wordpiece tokenization (Wu, et al., 2016; Schuster & Nakajima, 2012) so that the tokenization process is done based on sub-words. This technique certainly causes the OOV case to be much reduced, from 401 OOV words to zero OOV words, because when it finds that certain word does not exist in the vocabulary, it will search for the longest sub-word of that word that exist in the vocabulary (greedy first-longest-match algorithm). One of the misclassification cases that is caused by OOV words from previous study and is now predicted correctly can be seen in Table 7. There is a typo on the word "sncak" which should be "snack" and the word "sncak" is not in the training data.

Table 7. Example case #1

| Review | kamar tidur dan kamar mandi bersih, sncak lengkap (*The bedroom and bathroom is clean, snack is full*) |
|---|---|
| Prediction | ac_P1-neg |
| Ground-truth | ac_P1-none |

Misclassification that still occur are caused by mislabeled data and there are keywords for the aspect category or other sentiments contained in the review when in fact the phrases that contain these keywords have a neutral sentiment polarity. Hence, it cannot be associated with positive sentiment as well as negative sentiment. In this study, the neutral sentiment class was not taken into consideration, so the aspect category that has neutral sentiment polarity should not be detected. Example of such cases is described in Table 8.

Table 8. Example case #2

| Review | Kamar (#209) berbau apek ketika pertama kali masuk. Nampaknya sudah lama tidak ditempati dan AC lama tidak dinyalakan. (*The room (# 209) smelled musty when I first entered. It seems that it has not been occupied for a long time and the old air conditioner has not been turned on for so long*) |
|---|---|
| Prediction | linen_P1-none |
| Ground-truth | linen_P1-pos |

Other misclassification causes are the same as the cause of one of the misclassification cases in previous studies; there is a contradictory sentiment towards an aspect category in a sentence. When such cases occur, since our study focuses heavily on how to detect negative sentiments rather than detecting positive sentiments, the instance is labeled as having negative sentiments even though the keyword from positive sentiments is also exist in the review. Therefore, it is pretty difficult for the model to give correct prediction.

However, the BERT pretrained model used in this study is a multilingual model so that the words contained in the vocabulary come from various languages and many basic words from Indonesian are not in the vocabulary. This causes the results of tokenization are often not very good because it even split the basic words. In fact, the basic word should not be splitted, only the prefixes need to be separated from the basic word. Therefore, it is likely that the semantic information from words cannot be represented well enough with the BERT pretrained model so that the model still cannot fully determine the keywords that indicate the existence of certain aspect category or sentiments correctly which is why such misclassification cases still occur.

## VII. CONCLUSION

In this paper, we transform ABSA task from single sentence classification task into sentence-pair classification task by constructing auxiliary sentences based on work done by Sun, et al. (2019). We conduct comparative experiment to compare task-solving approach and pretrained model adaptation strategy used to solve ABSA task, perform hyperparameter tuning, and built the final model by fine-tuned the BERTpretrained model using sentence-pair Indonesian reviews dataset. We compare evaluation result with the result from previous study. We also analyze misclassification cases which still occur and discussed the vocabulary issue on using a multilingual pretrained model.

For future works, we suggest to train a new wordpiece vocabulary and pretrain BERT language model which is specific for Indonesian language to tackle the vocabulary issue in multilingual model. Try combining different types of text i.e. news, review, and academic for pretraining BERT so the model will be exposed to various text structures and contexts thus yields better language representation.

ACKNOWLEDGMENT

We would like to thank Airyrooms for providing us with the Indonesian hotel reviews dataset and for the cooperation during research period.